\documentclass[letterpaper]{article} 
\usepackage[submission]{aaai2026}  
\usepackage{times}  
\usepackage{helvet}  
\usepackage{courier}  
\usepackage[hyphens]{url}  
\usepackage{graphicx} 
\usepackage{natbib}  
\usepackage{caption} 
\usepackage{algorithm}
\usepackage{algpseudocode}
\usepackage{amsmath}
\usepackage{amsfonts}
\usepackage{tabularx}
\usepackage{booktabs}
\usepackage{multirow}

\usepackage{newfloat}
\usepackage{listings}
\DeclareCaptionStyle{ruled}{labelfont=normalfont,labelsep=colon,strut=off} 
\floatstyle{ruled}
\newfloat{listing}{tb}{lst}{}
\floatname{listing}{Listing}
\title{Exploring Weaknesses in Function Call Models via Reinforcement Learning: An Adversarial Data Augmentation Approach}

\author {
    Weiran Guo\textsuperscript{\rm 1},
    Bing Bo\textsuperscript{\rm 2},
    Shaoxiang Wu\textsuperscript{\rm 2},
    Jingsheng Yang\textsuperscript{\rm 2}
}
\affiliations {
    \textsuperscript{\rm 1}Department of Computer Science, Tongji University\\
    \textsuperscript{\rm 2}ByteDance\\
}

\usepackage{bibentry}

\begin{document}

\maketitle

\begin{abstract}
Function call capabilities have become crucial for Large Language Models (LLMs), enabling them to interact more effectively with external tools and APIs. Existing methods for improving the function call capabilities of LLMs rely on data obtained either through manual annotation or automated generation by models, and use this data to fine-tune the LLMs. However, these methods often lack targeted design and are constrained by fixed patterns and data distributions, which limits their effectiveness in enhancing the generalization and robustness of function call LLMs. To address this limitation, we propose a novel adversarial data augmentation method that employs reinforcement learning to systematically identify and target the weaknesses of function call LLMs. Our training framework introduces a query model trained with reinforcement learning (RL) to generate adversarial queries that are specifically designed to challenge function call (FC) models. This approach adopts a zero-sum game formulation, where the query model and the FC model engage in iterative alternating training. Overall, our method advances the development of more robust FC models and provides a systematic way to identify and correct weaknesses in the ability of LLMs to interact with external tools.
\end{abstract}


\section{Introduction}
\begin{figure}[htbp]  
  \centering
  \includegraphics[width=\columnwidth]{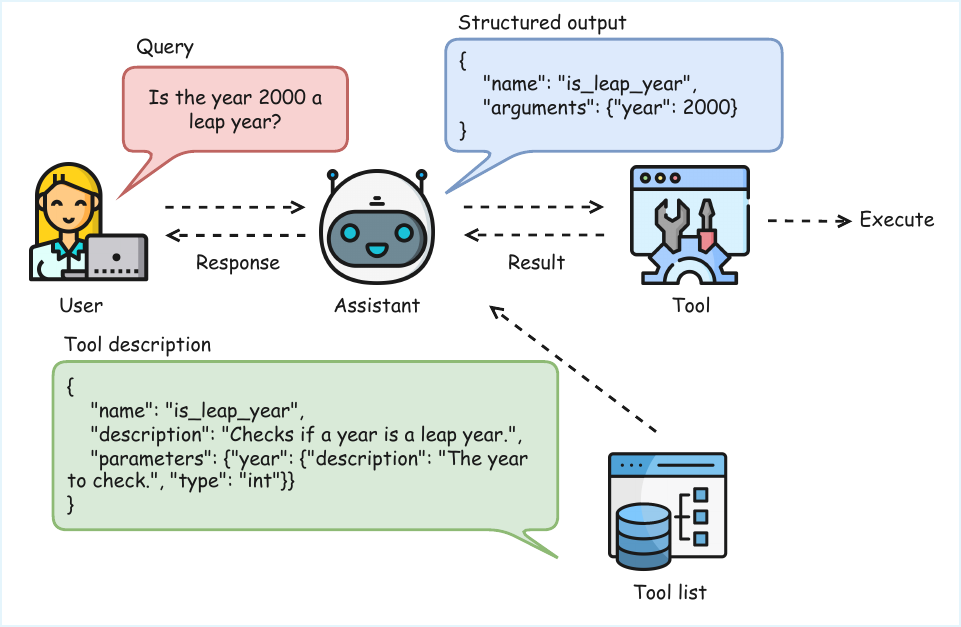}
  \caption{The illustration of a function calling process.}
  \label{fig:function_call_desc}
\end{figure}

Large Language Models (LLMs) such as GPT-4 \cite{openai2024gpt4technicalreport}, Qwen \cite{bai2023qwentechnicalreport}, and Gemini \cite{geminiteam2025geminifamilyhighlycapable} have shown strong general language capabilities. To extend their usefulness, many of these models now support function calling to interact with external tools and APIs. Function calling capabilities have emerged as a critical feature of modern LLMs, enabling them to interact with external tools, APIs, and services \cite{benchmark_fc, qin2024toolllm,wang2025function}. The function call process is illustrated in Figure \ref{fig:function_call_desc}. The user initiates a query, and the LLM, based on the available tool list and their descriptions, converts the query into a formatted input (e.g., in JSON format) that meets the requirements of the selected tool, including the chosen function and its corresponding parameter values. This formatted data is then passed to the designated tool for parsing and execution. After execution, the result is returned to the LLM, which then generates a response for the user. This capability forms the foundation for LLMs to perform complex tasks such as retrieving information, controlling applications, and executing operations on behalf of users. 

As LLMs increasingly serve as interfaces between users and digital systems, their ability to accurately interpret user requests and translate them into appropriate function calls becomes increasingly important for their practical utility. To further improve the function call performance of LLMs, existing work trains the base model using methods such as Supervised Fine-Tuning (SFT) \cite{wang2025toolgen} or Reinforcement Learning (RL) \cite{christiano_deep_2017,qian2025toolrlrewardtoollearning}. SFT trains LLMs on labeled input-output pairs to align their behavior with desired responses. RL, on the other hand, optimizes LLMs by rewarding desirable outputs, allowing them to improve through interaction and feedback.


Datasets used for SFT and RL training are typically created either through manual annotation by human experts or via automated labeling processes using LLMs \cite{liu2024apigen,ADC}. However, these datasets possess inherent limitations that affect their effectiveness. In the fine-tuning of LLMs, an important goal is to cover as many failure or bad cases of the Function Call (FC) model as possible to improve its robustness. Unfortunately, both manually annotated data and data generated by LLMs often lack the necessary specificity and are not deliberately designed to reveal the model’s weaknesses. Additionally, manually labeled datasets tend to follow limited and repetitive patterns, while automatically generated data from LLMs often suffers from fixed distributions and insufficient diversity. From the perspective of annotation efficiency, the manual creation of targeted error cases is not only labor-intensive but also time-consuming, making it difficult to scale and cover the broad range of potential failure modes effectively.

To address the above issues, we propose a new and more efficient training data augmentation method specifically designed for FC models. The core of our approach is to leverage RL to actively explore the bad cases of the FC model, generate highly targeted datasets that expose the model's weaknesses, and use these datasets to fine-tune the FC model. The goal is to reduce the bad case rate and increase the diversity of the training data. The contributions of this paper are as follows:

\begin{itemize}
    \item We train a query model using RL, leveraging its exploratory nature to thoroughly search for weaknesses in the FC model. By combining RL with a reward design based on zero-sum game theory and reasoning, the query model rewrites existing seed data to generate targeted fine-tuning data for the FC model, thereby achieving effective data augmentation.
    \item Using adversarial training, the query model and the FC model are trained in an alternating iterative manner. Specifically, the FC model is fine-tuned with data augmented by the query model to improve its accuracy and generalization. Then, bad cases are searched based on the trained FC model. In conjunction with curriculum learning, the difficulty and complexity of the FC data are gradually increased throughout the iterative process.
    \item During the training of the query model, we use an embedding loss to ensure the diversity of the rewritten data and set an early stopping criterion as the termination signal to maintain data quality.
\end{itemize}

\section{Related Work}
\label{sec:related_work}
\paragraph{Function Calling in LLMs}
Function calling has emerged as a key functionality in modern LLMs \cite{yao2023react,qin2024tool,patil2024gorilla}. This capability is essential for supporting complex tasks in real-world applications. Recent studies have proposed various methods to enhance its accuracy and robustness, using methods like token representation \cite{schick2023toolformer,wang2025toolgen} and multi-turn interaction \cite{lin2025robust}. In the training of FC models, data plays a crucial role. Early training relied on manually annotated data for fine-tuning FC models. With the advancement of LLMs, an increasing number of studies have started using LLMs to generate training data. APIGen \cite{liu2024apigen} leverages LLMs to automatically generate data through a Chain-of-Thought approach. In addition, another model is used for semantic verification to ensure higher-quality training data for the FC model. Similarly, ADC \cite{ADC} uses an automated approach to construct challenging datasets, and additionally incorporates coding tasks into training to enhance the FC model’s ability to follow format constraints. There are also many data construction methods targeting specific types of FC models, such as datasets designed for large tool catalogs, long tool responses, and long multi-turn conversations \cite{kate2025longfuncevalmeasuringeffectivenesslong}, as well as function call datasets covering diverse tool-use scenarios \cite{chen2025acebenchwinsmatchpoint} and data for complex tool calling \cite{zhong2025complexfuncbenchexploringmultistepconstrained}.
The above training methods have shown strong performance in improving FC models, but none of them consider generating targeted data based on the model's weaknesses.

\paragraph{Two-Player RL-based Training in LLMs}
Some existing works have extensively explored two-player LLM reinforcement learning (RL) training. \cite{zhou2024reflect} and \cite{ma2024coevolving} leverage the interaction or cooperation between two LLMs to perform RL-based fine-tuning. Recently, an increasing number of research works have adopted two-player RL for adversarial training. The core idea of adversarial training is to introduce challenging samples to the model, making it more robust and generalizable \cite{goodfellow2014explaining,goodfellow2014generativeadversarialnetworks}, while RL is a learning paradigm in which an agent learns in the form of rewards or penalties. In recent studies, RL-based adversarial training has been used to enhance the specific capabilities of large language models. In the field of LLM safety, DuoGuard \cite{deng2025duoguardtwoplayerrldrivenframework} and GPO \cite{zheng2024optimalllmalignmentsusing} leverage RL-based adversarial learning with an attacker-defender training framework to improve the accuracy of guardrail classification models and the response safety of LLMs, respectively. In addition to adversarial training using two different models, there are also some works that adopt self-play with a single model \cite{cheng2024selfplaying,self-fine,wu2025selfplay}. Absolute Zero \cite{zhao2025absolutezeroreinforcedselfplay} adopts a self-play approach, enabling the model to engage in self-generated coding tasks by simultaneously proposing and solving problems in a self-adversarial manner, thereby achieving zero-shot training. These methods have not been validated in the function calling domain, and they rely on simultaneous training of the attacker and defender, which lacks stability and controllability.

\section{Priliminaries}
\paragraph{Problem Setting}
An LLM can be represented as a policy $\pi_{\theta}$ parameterized by $\theta$. Given a prompt $\mathbf{x}$, the LLM generates a response $\mathbf{y}=(y_1,y_2,\dots,y_T)$ in an autoregressive manner, producing token-level outputs, where $y_t$ is the $t$-th token of $y$. Hence, the distribution of $\pi_{\theta}$ can be represented as
\begin{equation}
    \pi_{\theta}(\mathbf{y}\mid\mathbf{x})=\prod_{t=1}^T\pi_{\theta}(y_t\mid \mathbf{x},y_1,y_2,\dots,y_{t-1}).
\end{equation}

In this paper, we construct an additional query model designed to explore and uncover the weaknesses of the FC model, specifically its bad cases. Both the query model and the FC model are LLMs, denoted as $\pi_Q$ and $\pi_F$, respectively. For simplicity, we omit the parameter $\theta$ when denoting the policies of the two models.

\paragraph{LLM Post-Training}
In the post-training phase of an LLM, a dataset $\mathcal{D}$ is constructed from pairs of input prompts $\mathbf{x}$ and their corresponding ground truth responses $\hat{\mathbf{y}}$, where each pair represents a single training instance. \textit{SFT} serves as a critical step in aligning the LLM's behavior with desired outputs by employing a behavior cloning–like strategy. Specifically, SFT utilizes high-quality, human-annotated data to explicitly shape the LLM's responses, encouraging the model to imitate the ground truth outputs observed in the dataset. The training objective in SFT is to minimize the discrepancy between the model’s predicted response and the corresponding ground truth, typically formulated as the following loss function:
\begin{equation}
    \mathcal{L}_{\text{SFT}}(\theta) = -\mathbb{E}_{(\mathbf{x}, \hat{\mathbf{y}}) \sim \mathcal{D}} \left[\log \pi_{\theta}(\hat{\mathbf{y}}\mid\mathbf{x}) \right].
\end{equation}
\textit{RL} guides the LLM to produce desired responses by leveraging the principles of exploration and exploitation, combined with a reward-based feedback mechanism. During training, the model is encouraged to explore different output strategies while exploiting known successful ones, receiving positive rewards for correct or high-quality outputs and negative rewards (or penalties) for incorrect or suboptimal responses. This process enables the LLM to learn policies that generalize beyond supervised examples. The objective of RL optimization is to maximize the expected cumulative reward, formalized as the following objective function:
\begin{equation}\label{rl_obj}
    \mathcal{J}_{\text{RL}}(\theta) = \mathbb{E}_{(\mathbf{x}, \hat{\mathbf{y}}) \sim \mathcal{D}, \mathbf{y} \sim \pi(\cdot \mid \mathbf{x})}\left[R(\mathbf{x}, \mathbf{y}, \hat{\mathbf{y}})\right].
\end{equation}
The reward function $R(\mathbf{x}, \mathbf{y}, \hat{\mathbf{y}})$ measures the quality of $\mathbf{y}$ and can be implemented through human feedback, models, or certain rules.

\paragraph{Zero-Sum Game}
A zero-sum game refers to a competitive scenario involving two players, in which one player’s gain is exactly balanced by the other player’s loss, resulting in a net payoff of zero. This type of interaction is commonly found in adversarial settings, where cooperation between the players offers no benefit, and any advantage gained by one party necessarily comes at the expense of the other. In the context of this paper, we model the interaction between the query model and the FC model as a zero-sum game. The query model is trained to deliberately generate challenging queries that expose the weaknesses of the FC model, thereby reducing the FC model’s performance. Conversely, the FC model is optimized to handle such adversarial queries correctly and maintain robust function call capabilities. Since the improvement of one model inherently hinders the other, the two models are locked in a competitive and adversarial dynamic. This setup naturally satisfies the conditions of a zero-sum game and forms the foundation for our iterative adversarial training framework.


\begin{figure*}[htbp]  
  \centering
  \includegraphics[width=\textwidth]{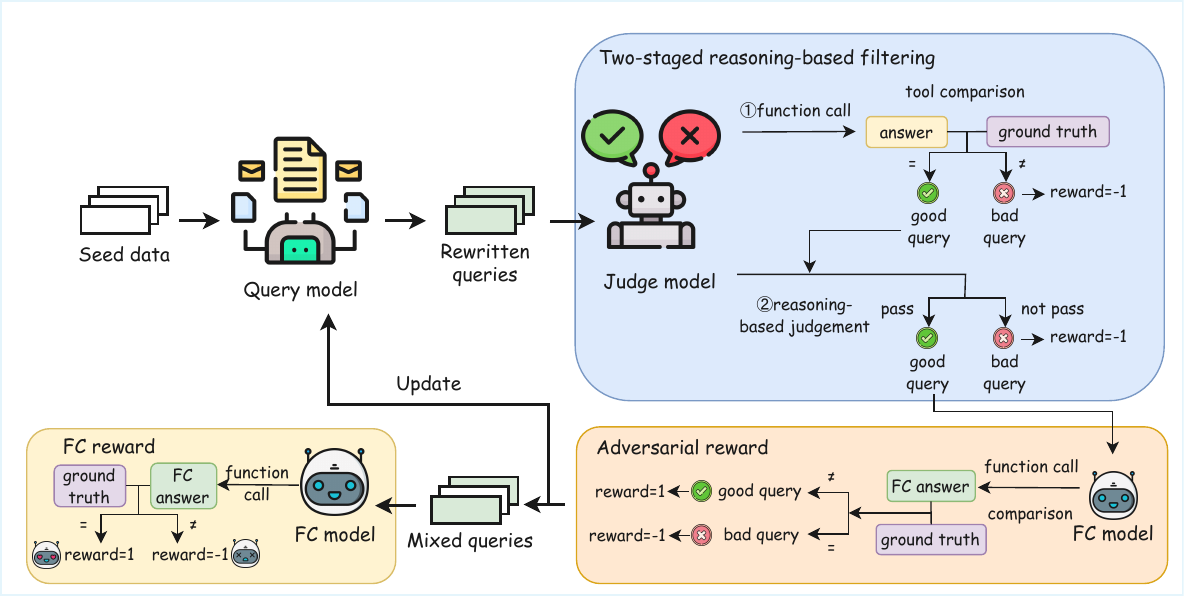}
  \caption{The complete process of one iteration of training.}
  \label{fig:training_process}
\end{figure*}

\section{Bad Case Exploring: Query Model}

\subsection{Data Preparation}
The training objective of the query model $\pi_Q$ is to use RL to actively explore queries or patterns of queries that are likely to cause the FC model to produce incorrect answers. This forms the foundation for training the FC model in this paper, as only by generating a sufficient number of bad cases as training data can we further improve the performance of the FC model.

Instead of having the $\pi_Q$ generate challenging queries from scratch, we adopt a query rewriting approach. We use a seed dataset $\mathcal{D}_\mathrm{seed}=\{(\mathbf{x}, \hat{\mathbf{y}})\}^{N}$ of a certain scale $N$, where each data point consists of an input query along with its context and the correct response that the assistant, i.e., the FC model, is expected to produce. We adopt the query rewriting approach over directly generating challenging queries from scratch, based on the following considerations:
(1) The queries in the seed dataset offer rich prompt diversity, which facilitates the training of $\pi_Q$ by exposing it to a wide range of input patterns.
(2) The seed dataset provides ground-truth responses, which serve as supervision signals to constrain and guide $\pi_Q$ during training.
(3) Rewriting also enables better control over the distribution of data types in the FC model’s training set, preventing $\pi_Q$ from focusing disproportionately on a single occasion (e.g., irrelevance-related samples). We denote a data point of the training set for $\pi_Q$ as $\mathbf{x}_Q=\mathrm{template}(\mathbf{x}, \hat{\mathbf{y}})$, where $\mathrm{template}(\cdot)$ is a function that transforms the data from the seed dataset into prompts for $\pi_Q$. 

\subsection{Reward Design}
To evaluate the quality of the rewritten queries, we need to design an appropriate reward for $\pi_Q$. The reward design for $\pi_Q$ in this paper follows two principles: ensuring the plausibility of the rewritten query and inducing the FC model to produce incorrect answers. To this end, we design the reward process as two sequential components: a two-stage filtering mechanism based on a reasoning-based LLM and an adversarial reward based on a zero-sum game. It is important to note that the following rewards are outcome-based, meaning each reward corresponds to a complete generation.

\paragraph{Two-stage Filtering Mechanism} We use a larger model as the judge model to perform two rounds of filtering on the queries rewritten by $\pi_Q$, assigning penalties to those that do not meet the quality criteria from the perspective of rewriting quality. 
\begin{enumerate}
    \item In the first stage, we input the rewritten query into the judge model and let it perform a function call based on the rewritten query. We compare the tool name $f_{\mathbf{y}'}$ in the answer $\mathbf{y}'$ generated by the judge model with the tool name $f_{\hat{\mathbf{y}}}$ in the ground truth answer $\hat{\mathbf{y}}$ to check whether they are the same. The reason for this is to preliminarily filter out rewritten queries that cannot be traced back to the original tool. If a rewritten query cannot be mapped back to the original tool, it indicates that the query has deviated from the semantics of the original one. We compare only the tool name because parameters can be complex with multiple correct answers. Focusing on the tool name allows for more flexibility in query rewriting and helps uncover a wider range of potential bad cases for the FC model.
    \item In the second stage, we leverage the judge model's reasoning capability to further evaluate the rewritten queries that passed the previous filtering step. In this stage of evaluation, we impose additional semantic constraints on the rewritten queries, such as whether key fields are missing or whether the query is posed from the perspective of a user. The reason for this design is that the first-stage evaluation does not involve reasoning and may result in judgement errors. The query model may occasionally generate rewritten queries that pass the first stage but still deviate from the original query’s semantics—for example, posing the query from the assistant’s perspective to mislead the FC model into acting as the user. We use a large-scale LLM to identify such rewritten queries and assign penalties accordingly. We define a boolean variable $\phi_\mathrm{valid}$ to represent the result of the second-stage evaluation. If $\phi_\mathrm{valid} = \text{True}$, it indicates that the rewritten query satisfies the semantic requirements.

\end{enumerate}
We adopt a combination of non-thinking and thinking evaluations because using only non-thinking judgements results in relatively low accuracy, while relying solely on thinking evaluations is time-consuming. The two-stage approach balances accuracy and efficiency. We use $r_\mathrm{judge}$ to denote the two-stage judgement reward, which can be expressed as:
\begin{equation}\label{judge_reward}
    r_{\mathrm{judge}} =
\begin{cases}
1,  & \text{if } f_{\mathbf{y}'} = f_{\hat{\mathbf{y}}} \;\land\; \phi_{\mathrm{valid}} = \text{True} \\
-1, & \text{otherwise}
\end{cases}
\end{equation}

\paragraph{Adversarial Reward}
Given that $\pi_Q$ and $\pi_F$ form a zero-sum game and are in an adversarial relationship, if $\pi_F$ produces an incorrect answer when performing the function call task based on the query rewritten by $\pi_Q$, it indicates that $\pi_Q$ has successfully generated a challenging query. In this case, $\pi_Q$ is rewarded; otherwise, it is penalized. We define this adversarial reward as $r_{\mathrm{adv}}$, and express it as:
\begin{equation}\label{adv_reward}
        r_{\mathrm{adv}} =
\begin{cases}
1,  & \text{if } r_\mathrm{judge} = 1 \;\land\;\mathbf{y} \neq \hat{\mathbf{y}}\\
-1, & \text{otherwise}
\end{cases}
\end{equation}
The above analysis indicates that when the output of $\pi_F$ deviates from the ground truth answer $\hat{\mathbf{y}}$, it implies that $\pi_Q$ has successfully generated an adversarially effective rewritten query $\Tilde{\mathbf{x}}$. In such cases, we regard the rewritten query as a successful attack and assign a positive reward to $\pi_Q$ to reinforce this behavior. Since the two reward components above are sequentially related, the adversarial reward $r_{\mathrm{adv}}$ takes $r_\mathrm{judge}$ into account.

Combining the results of the two sequential reward assignments above, we obtain the final reward for $\pi_Q$ as $R_Q(\mathbf{x},\Tilde{\mathbf{x}},\hat{\mathbf{y}}) = r_{\mathrm{adv}}$.

\subsection{Diversity: Embedding Loss}
Due to the inherently exploitative nature of RL, LLMs tend to converge toward generating highly similar or even repetitive outputs during the later stages of training. This behavior, while potentially beneficial for optimizing a specific reward signal, can be detrimental in scenarios where output diversity is important. In our case, we aim to construct a high-quality training dataset for the follow-up classification model $\pi_F$, which relies on diverse and informative rewritten queries generated by the query rewriting model $\pi_Q$. If $\pi_Q$ produces overly similar rewrites, the resulting dataset may lack sufficient variability, which in turn can limit the generalization ability of $\pi_F$.

To address this issue, we propose a method to explicitly encourage diversity in the outputs of $\pi_Q$. Specifically, we introduce an embedding-based regularization term into the training objective of $\pi_Q$. This regularization term is designed to promote higher diversity within each training batch by encouraging larger edit distances between different rewritten queries in the batch. By doing so, we aim to prevent entropy collapse and ensure that $\pi_Q$ explores a broader range of possible rewrites while still maintaining semantic fidelity to the original query. The regularized objective function for training $\pi_Q$ is defined as follows:
\begin{align}
\mathcal{J}_{Q}(\theta) 
&= \mathbb{E}\left[R_Q(\mathbf{x}, \Tilde{\mathbf{x}}, \hat{\mathbf{y}})\right] \notag \\
&\quad + \alpha \left( \frac{2}{B(B - 1)} \sum_{i=1}^{B} \sum_{j=i+1}^{B} \left(1 - \frac{\mathbf{v}_i \cdot \mathbf{v}_j}{\|\mathbf{v}_i\| \|\mathbf{v}_j\|} \right) \right),
\end{align}

where $\alpha$ is a parameter that controls the weight of the regularization term, $B$ is the batch size, and $\mathbf{v}_i$ and $\mathbf{v}_j$ are the embeddings of the $i$-th and $j$-th rewritten queries in a batch.

\subsection{Early Stop}
In RL training, LLMs typically converge to a stable reward, at which point they tend to generate similar patterns in response to different original queries. However, our goal is to obtain diverse rewriting patterns for the queries. To achieve this, we leverage the exploratory nature of RL by collecting all the bad cases of the FC model encountered during sampling in training, and we terminate the training process early at a certain timestep. Training is terminated when the reward meets the condition
$\left| \mathbb{E}\left[R_{Q,\tau}(\mathbf{x},\Tilde{\mathbf{x}},\hat{\mathbf{y}})\right]-\mathbb{E}\left[R_{Q,\tau-1}(\mathbf{x},\Tilde{\mathbf{x}},\hat{\mathbf{y}})\right]\right|<\epsilon$ within 10 timesteps,
where $\tau$ denotes the timestep, and $\epsilon$ is the threshold for early stopping, which is manually determined. This indicates that we stop training and query collection once $\pi_Q$ has converged to a certain extent.

\section{FC Model Post-Training}
A rewritten query $\Tilde{\mathbf{x}}$ generated by $\pi_Q$, paired with the ground truth output corresponding to the original query, constitutes a single data point $(\Tilde{\mathbf{x}}, \hat{\mathbf{y}})$ in the rewritten dataset. The rewritten dataset is used as the training set for the FC model $\pi_F$. We use SFT (full fine-tuning and LoRA) and RL to fine-tune $\pi_F$. For the RL training of $\pi_F$, we compare the $\pi_F$’s output $\mathbf{y}$ with the ground truth answer $\hat{\mathbf{y}}$ to obtain the following reward design:
\begin{equation}\label{fc_reward}
        R_{\mathrm{F}}(\Tilde{\mathbf{x}},\mathbf{y},\hat{\mathbf{y}}) =
\begin{cases}
1,  & \text{if } \mathbf{y} = \hat{\mathbf{y}}\\
-1, & \text{otherwise}
\end{cases}
\end{equation}
Compared with Equation (\ref{adv_reward}), the adversarial relationship between $\pi_Q$ and $\pi_F$ becomes clear.

\section{Iterative Alternating Training With Curriculum Learning}
The FC model, after undergoing one round of fine-tuning, can participate again in the training of the query model, forming an iterative training process between the two. One iteration consists of training the query model once and the FC model once. The FC model trained in the $k$‑th iteration is denoted as $\pi_{F,k}$, respectively. The dataset used to train $\pi_{Q}$ in the $k$‑th iteration is $\mathbb{D}_k$, and the training set for $\pi_{F,k}$, generated by $\pi_{Q}$, is $\mathcal{D}_k$.

In order to help the follow-up FC model gradually acquire the ability to handle increasingly complex types of input data, we adopt a progressive learning strategy by adjusting the composition of the training dataset $\mathcal{D}_k$ at each stage of the training process. This approach enables the model to build a strong foundation on simpler examples before being exposed to more difficult and realistic scenarios, thereby improving its generalization and robustness. For example, in early iterations of training, we focus on cases where a query corresponds to a single function call, and in the later iterations, on cases with multiple function calls per query. The algorithmic process of the iterative training is shown in Algorithm \ref{alg}. The complete process of one iteration of training of our method is shown in Figure \ref{fig:training_process}.

\begin{table*}[htbp]
\centering
\begin{tabular}{@{}lcccccc@{}}
\toprule
Model &
  round &
  \begin{tabular}[c]{@{}c@{}}Overall\\ Acc\end{tabular} &
  \begin{tabular}[c]{@{}c@{}}Non-Live\\ AST Acc\end{tabular} &
  \begin{tabular}[c]{@{}c@{}}Live\\ Acc\end{tabular} &
  \begin{tabular}[c]{@{}c@{}}Relevance\\ Detection\end{tabular} &
  \begin{tabular}[c]{@{}c@{}}Irrelevance\\ Detection\end{tabular} \\ \midrule
Qwen3-0.6B                                &   & 39.27\%          & 56.73\%          & 53.93\%          & 66.67\%          & \textbf{90.35\%} \\
\multirow{2}{*}{Qwen3-0-6B (LLM + LoRA)} & 1 & 39.26\%          & 52.73\%          & 59.22\%          & 88.89\%          & 76.34\%          \\
                                          & 2 & 37.82\%          & 48.58\%          & 58.60\%          & {\underline{83.33\%}}    & 75.20\%          \\
\multirow{2}{*}{Qwen3-0.6B (ours + LoRA)} & 1 & 41.77\%          & 62.31\%          & 59.53\%          & 94.44\%          & 74.03\%          \\
                                          & 2 & 39.41\%          & 52.42\%          & 60.02\%          & \textbf{88.89\%} & 76.53\%          \\
\multirow{2}{*}{Qwen3-0.6B (LLM + SFT)}   & 1 & 32.15\%          & 32.27\%          & 56.51\%          & 77.78\%          & 74.85\%          \\
                                          & 2 & 33.59\%          & 38.56\%          & 55.00\%          & 61.11\%          & {\underline{78.16\%}}    \\
\multirow{2}{*}{Qwen3-0.6B (ours + SFT)}  & 1 & 31.41\%          & 36.17\%          & 54.51\%          & 83.33\%          & 62.44\%          \\
                                          & 2 & 34.94\%          & 55.00\%          & 50.64\%          & \textbf{88.89\%} & 54.61\%          \\
\multirow{2}{*}{Qwen3-0.6B (LLM + RL)}    & 1 & 41.23\%          & 61.02\%          & 60.20\%          & 88.89\%          & 70.06\%          \\
                                          & 2 & {\underline{42.60\%}}    & {\underline{64.31\%}}    & {\underline{61.35\%}}    & {\underline{83.33\%}}    & 72.36\%          \\
\multirow{2}{*}{Qwen3-0.6B (ours + RL)}   & 1 & 39.09\%          & 53.58\%          & 59.53\%          & 88.89\%          & 69.75\%          \\
                                          & 2 & \textbf{44.21\%} & \textbf{69.10\%} & \textbf{61.84\%} & 77.78\%          & 75.09\%          \\ \bottomrule
\end{tabular}
\caption{Performance comparison of our method with the baseline and the base model on Qwen3-0.6B. The fine-tuned models are denoted in the format: \textless{}base model name\textgreater (\textless{}data augmentation method\textgreater + \textless{}fine-tuning method\textgreater). Bold numbers indicate the highest metric values in the final training round, while underlined numbers represent the second-highest values. The same notation applies below.}
\label{tab:efficiency}
\end{table*}

\begin{table*}[tb]
\centering
\begin{tabular}{@{}lllllll@{}}
\toprule
Model &
  round &
  \begin{tabular}[c]{@{}l@{}}Overall\\ Acc\end{tabular} &
  \begin{tabular}[c]{@{}l@{}}Non-Live\\ AST Acc\end{tabular} &
  \begin{tabular}[c]{@{}l@{}}Live\\ Acc\end{tabular} &
  \begin{tabular}[c]{@{}l@{}}Relevance\\ Detection\end{tabular} &
  \begin{tabular}[c]{@{}l@{}}Irrelevance\\ Detection\end{tabular} \\ \midrule
Qwen3-8B                                &   & \textbf{55.64\%} & \textbf{88.90\%} & {\underline{73.88\%}}    & \textbf{94.44\%}  & \textbf{71.42\%} \\
\multirow{2}{*}{Qwen3-8B (LLM + RL)}    & 1 & 53.74\%          & 85.90\%          & \textbf{75.66\%} & 83.33\%           & 76.44\%          \\
                                        & 2 & {\underline{52.96\%}}    & {\underline{88.02\%}}    & 72.63\%          & {\underline{88.89\%}}     & 70.02\%          \\
\multirow{2}{*}{Qwen3-8B (ours + RL)}   & 1 & 46.15\%          & 60.98\%          & 72.50\%          & 88.89\%           & 75.79\%          \\
                                        & 2 & 52.89\%          & 87.52\%          & 72.99\%          & {\underline{88.89\%}}     & 69.78\%          \\ \midrule
Qwen3-4B                                &   & {\underline{52.18\%}}    & \textbf{87.33\%} & 68.41\%          & \textbf{100.00\%} & 62.21\%          \\
\multirow{2}{*}{Qwen3-4B (LLM + RL)}    & 1 & 44.24\%          & 48.79\%          & 70.01\%          & 83.33\%           & 70.97\%          \\
                                        & 2 & 50.96\%          & 84.56\%          & {\underline{69.57\%}}    & {\underline{94.44\%}}     & {\underline{69.84\%}}    \\
\multirow{2}{*}{Qwen3-4B (ours + RL)}   & 1 & 51.51\%          & 69.67\%          & 75.43\%          & 88.89\%           & 79.23\%          \\
                                        & 2 & \textbf{52.88\%} & {\underline{86.15\%}}    & \textbf{72.90\%} & 88.89\%           & \textbf{72.81\%} \\ \midrule
Qwen2.5-3B-Instruct            &   & {\underline{43.82\%}}    & {\underline{78.02\%}}    & 55.89\%          & \textbf{94.44\%}  & 55.59\%          \\
\multirow{2}{*}{Qwen2-5-3B (LLM + RL)}  & 1 & 42.35\%          & 55.38\%          & 66.37\%          & 77.78\%           & 70.95\%          \\
                                        & 2 & 47.26\%          & 67.75\%          & \textbf{70.32\%} & 72.22\%           & \textbf{78.10\%} \\
\multirow{2}{*}{Qwen2-5-3B (ours + RL)} & 1 & 38.74\%          & 41.46\%          & 66.64\%          & 72.22\%           & 72.62\%          \\
                                        & 2 & \textbf{49.87\%} & \textbf{80.38\%} & {\underline{69.21\%}}    & \textbf{94.44\%}  & {\underline{71.73\%}}    \\ \bottomrule
\end{tabular}
\caption{Performance comparison of our method across language models of different sizes.}
\label{tab:multi-scale}
\end{table*}

\begin{algorithm}[ht]
\caption{}\label{alg}
\begin{algorithmic}[1]
\Require The base model of the query model $\pi_{Q}$; initial FC model $\pi_{F,0}$; seed data $\mathcal{D}_\mathrm{seed}$; batch size $B$
\For{$k = 0, 1, 2, \dots$}
    \State Construct training data $\mathbb{D}_k$ for $\pi_{Q}$ based on $\mathcal{D}_\mathrm{seed}$
    \State Initialize the dataset $\mathcal{D}_k$
    \For{$\tau = 0, 1, 2, \dots$}
        \State $\pi_{Q}$ generates rewritten queries $\Tilde{\mathbf{X}}=(\Tilde{\mathbf{x}}_1, \Tilde{\mathbf{x}}_2, \dots, \Tilde{\mathbf{x}}_B)$ based on $\mathbb{D}_k$
        \State The judge model assesses the semantic quality of the rewritten queries in $\Tilde{\mathbf{X}}$ according to Equation (\ref{judge_reward})
        \State $\pi_{F,k}$ generates ${\mathbf{Y}}=({\mathbf{y}}_1, {\mathbf{y}}_2, \dots, {\mathbf{y}}_B)$ based on $\Tilde{\mathbf{X}}$
        \State Obtain the reward of $\pi_{Q}$ according to Equation (\ref{adv_reward})
        \State Collect the rewritten queries with negative rewards and the corresponding ground truth into $\mathcal{D}_k$
        \State Update $\pi_{Q}$ according to Equation (\ref{rl_obj})
    \EndFor
    \For{$\tau = 0, 1, 2, \dots$}
        \State $\pi_{F,k}$ generates ${\mathbf{Y}'}=({\mathbf{y}}'_1, {\mathbf{y}}'_2, \dots, {\mathbf{y}}'_B)$ based on $\mathcal{D}_k$
        \State Obtain the reward of $\pi_{F,k}$ according to Equation (\ref{fc_reward})
        \State Update $\pi_{F,k}$ according to Equation (\ref{rl_obj})
    \EndFor
    \State Obtain the updated FC model $\pi_{F,k+1}$
\EndFor
\end{algorithmic}
\end{algorithm}

\section{Experiments}
We use a high-quality internal dataset manually annotated based on real tools, which closely reflects real-world scenarios and validates the practical effectiveness of our method. Although this dataset is used as seed data, our approach is independent of the specific seed choice and remains broadly applicable.

We conduct two rounds of iterative adversarial training. In the first round, about 1,700 internal data samples are used as seed data. In the second round, 300 parallel-type samples from xlam-function-calling-60k \cite{liu2024apigen} are added. The query model is based on Qwen2.5-7B-Instruct, while the judge model is Qwen3-32B. For computing sentence embeddings, we use the text2vec-base-multilingual model \cite{text2vec}, as the queries involve both Chinese and English. We set the coefficient $\alpha$ of the embedding regularization term to 0.05 and the early stopping threshold $\epsilon$ to 0.2 based on empirical observations. Our code is based on the mainstream LLM reinforcement learning fine-tuning framework, verl \cite{sheng2024hybridflow}. For the RL training of both the query model and the FC model, we use the mainstream RL algorithm PPO \cite{schulman2017proximalpolicyoptimizationalgorithms}. We use NVIDIA H20 GPUs uniformly for all training. The query models are trained on 8 GPUs; experiments with Qwen3-0.6B and Qwen3-1.7B use 4 GPUs; Qwen3-4B is trained on 8 GPUs; and Qwen3-8B use 16 GPUs.

To validate the effectiveness of our approach, we adopt the Berkeley Function-Calling Leaderboard (BFCL) \cite{patil2025bfcl} as our evaluation framework. The BFCL evaluation set includes various data metrics. We focus on the performance of the FC model fine-tuned with our augmented data in four aspects: non-live, live, relevance detection, and irrelevance detection. Considering latency concerns in function call scenarios, all our experiments are conducted in the non-thinking mode. To align with our training data, we evaluate the base model using the prompt mode.  

\subsection{Effectiveness Verification}
To demonstrate the effectiveness of our approach, we use Qwen3-0.6B as the base model and investigate the advantages of our method compared to directly using an LLM to sample and generate training data. For the query rewriting component, we generate rewritten queries using both a trained query model and LLM sampling. To ensure a consistent data volume between the two methods, we randomly sample approximately 5,100 rewritten queries from those generated by the query model based on the seed data in each round. Similarly, the LLM is used to repeatedly sample from the seed data to produce another set of around 5,100 rewritten queries. These two sets are then respectively used as training data for the FC model. For fine-tuning the FC model, we combine the rewritten queries with the seed data as the training set, and perform training using three approaches: full-parameter SFT, LoRA, and RL. The results of this experiment are shown in Table \ref{tab:efficiency}. We evaluate the effectiveness of our method in data augmentation from multiple perspectives. 

From a side-by-side comparison, we compare our method against the baseline and the base model. Clearly, fine-tuning the FC model with datasets augmented by our method results in higher overall function call accuracy than using datasets augmented directly through LLM sampling. Meanwhile, for different sub-metrics, the FC models trained with our augmented datasets using RL achieve higher accuracy on both the non-live and live evaluation sets compared to the baseline and base model. Additionally, FC models fine-tuned with our datasets using full-parameter SFT and LoRA show higher accuracy in relevance detection. However, for the irrelevance detection metric, the base model still performs best. This is because, without undergoing function call training, the base model tends to reject a wide range of queries.

From a longitudinal perspective, observing the changes in our method after multiple rounds of training, we find that the model's performance on various function call metrics slightly declines after the first round. This is because the model has not yet fully fitted the distribution of the dataset. After the second round, the models fine-tuned with full-parameter SFT and RL show a clear improvement in overall performance, indicating that the second round of training significantly enhances the model’s capabilities.

Among the three fine-tuning methods for the FC model, full-parameter SFT is the least stable and may even result in worse performance than the base model. This is because full-parameter SFT is prone to overfitting. Between LoRA and RL fine-tuning, LoRA shows limited improvement over the base model, while RL consistently achieves the most significant performance gains on Qwen3-0.6B, both with the baseline dataset and our augmented dataset.

\subsection{Experiments on Multiple Base Models}
To explore the performance of our method across models of different scales, we select Qwen2.5-3B-Instruct, Qwen3-0.6B (which can be seen in the previous experiment), Qwen3-4B, and Qwen3-8B. We use RL—which showed relatively stable results in the previous experiment—as the training approach for the FC model. All other configurations remain the same as in the previous experiment.

First, we compare the function call accuracy trends of our data augmentation method across language models of different scales. Our method shows more notable improvements over the baseline and base model on smaller models such as Qwen3-0.6B, Qwen3-4B, and Qwen2.5-7B-Instruct. Specifically, for Qwen3-0.6B, the model trained on our augmented dataset achieves an accuracy that is 1.61\% higher than the baseline and 4.94\% higher than the base model. For Qwen2.5-7B-Instruct, our method improves accuracy by 2.61\% over the baseline and 6.05\% over the base model. However, for a larger model like Qwen3-4B, the improvements shrink to 1.92\% and 0.7\% over the baseline and base model, respectively, indicating a narrowing performance gap. For larger models like Qwen3-8B, our method does not show a noticeable advantage over the baseline and base model. We identify three main reasons for this: First, Qwen3-8B already possesses strong function call capabilities, which increases the proportion of low-quality rewritten queries that exploit reward hacking, making it difficult to identify high-quality bad cases and resulting in suboptimal FC model training. Second, the query rewriting ability is limited by the size of the query model, and training larger FC models requires correspondingly larger query models. Third, as the scale of the FC model increases, stricter judgment is required from the judge model. When using Qwen3-32B as the judge for a large FC model, reward hacking becomes more likely. The above experimental results show that our method yields more significant improvements on smaller models, while the gains on larger models are less pronounced. This suggests that our approach is particularly suitable for scenarios where low latency or offline deployment is required, such as chatbots and mobile devices.


By observing the accuracy trend of the FC model across the training iterations, we arrive at a conclusion consistent with the findings of the previous experiment. Specifically, after the first round of RL training, the models trained using our augmented dataset exhibit a noticeable decline in performance across several evaluation metrics. Following the second round of training, we observe a substantial improvement in the model’s performance, with marked gains in function call accuracy and other sub-metrics. These results highlight the importance of multi-stage training when leveraging complex data augmentation strategies.

\section{Conclusion}
In this work, we present a novel adversarial data augmentation framework to enhance the function call capabilities of LLMs. By formulating the training process as a zero-sum game between a reinforcement learning-based query model and an FC model, our approach systematically exposes and addresses the weaknesses of FC models. Unlike existing methods that rely on static or manually constructed data, our method dynamically generates challenging adversarial queries tailored to the vulnerabilities of the FC model. Experimental results demonstrate that our approach improves both the robustness and generalization of LLMs in function call tasks, offering a principled and effective path forward for aligning LLMs with real-world tool-use demands. In addition to the function call task, the method proposed in this paper can also be applied to other areas and scenarios of large language models (LLMs). In future work, we will explore more optimization methods related to function calls and attempt to apply them to various scenarios beyond function calls.

\bibliography{aaai2026}

@misc{openai2024gpt4technicalreport,
      title={GPT-4 Technical Report}, 
      author={OpenAI},
      year={2024},
      eprint={2303.08774},
      archivePrefix={arXiv},
      primaryClass={cs.CL},
      url={https://arxiv.org/abs/2303.08774}, 
}

@misc{bai2023qwentechnicalreport,
      title={Qwen Technical Report}, 
      author={Jinze Bai and Shuai Bai and Yunfei Chu and Zeyu Cui and Kai Dang and Xiaodong Deng and Yang Fan and Wenbin Ge and Yu Han and Fei Huang and Binyuan Hui and Luo Ji and Mei Li and Junyang Lin and Runji Lin and Dayiheng Liu and Gao Liu and Chengqiang Lu and Keming Lu and Jianxin Ma and Rui Men and Xingzhang Ren and Xuancheng Ren and Chuanqi Tan and Sinan Tan and Jianhong Tu and Peng Wang and Shijie Wang and Wei Wang and Shengguang Wu and Benfeng Xu and Jin Xu and An Yang and Hao Yang and Jian Yang and Shusheng Yang and Yang Yao and Bowen Yu and Hongyi Yuan and Zheng Yuan and Jianwei Zhang and Xingxuan Zhang and Yichang Zhang and Zhenru Zhang and Chang Zhou and Jingren Zhou and Xiaohuan Zhou and Tianhang Zhu},
      year={2023},
      eprint={2309.16609},
      archivePrefix={arXiv},
      primaryClass={cs.CL},
      url={https://arxiv.org/abs/2309.16609}, 
}

@misc{geminiteam2025geminifamilyhighlycapable,
      title={Gemini: A Family of Highly Capable Multimodal Models}, 
      author={Gemini Team},
      year={2025},
      eprint={2312.11805},
      archivePrefix={arXiv},
      primaryClass={cs.CL},
      url={https://arxiv.org/abs/2312.11805}, 
}

@inproceedings{benchmark_fc,
  author={Shijue Huang and Wanjun Zhong and Jianqiao Lu and Qi Zhu and Jiahui Gao and Weiwen Liu and Yutai Hou and Xingshan Zeng and Yasheng Wang and Lifeng Shang and Xin Jiang and Ruifeng Xu and Qun Liu},
  title={Planning, Creation, Usage: Benchmarking LLMs for Comprehensive Tool Utilization in Real-World Complex Scenarios},
  year={2024},
  cdate={1704067200000},
  pages={4363-4400},
  url={https://doi.org/10.18653/v1/2024.findings-acl.259},
  booktitle={ACL (Findings)}
}

@inproceedings{qin2024toolllm,
title={Tool{LLM}: Facilitating Large Language Models to Master 16000+ Real-world {API}s},
author={Yujia Qin and Shihao Liang and Yining Ye and Kunlun Zhu and Lan Yan and Yaxi Lu and Yankai Lin and Xin Cong and Xiangru Tang and Bill Qian and Sihan Zhao and Lauren Hong and Runchu Tian and Ruobing Xie and Jie Zhou and Mark Gerstein and dahai li and Zhiyuan Liu and Maosong Sun},
booktitle={The Twelfth International Conference on Learning Representations},
year={2024},
url={https://openreview.net/forum?id=dHng2O0Jjr}
}

@inproceedings{
liu2024apigen,
title={{APIG}en: Automated {PI}peline for Generating Verifiable and Diverse Function-Calling Datasets},
author={Zuxin Liu and Thai Quoc Hoang and Jianguo Zhang and Ming Zhu and Tian Lan and Shirley Kokane and Juntao Tan and Weiran Yao and Zhiwei Liu and Yihao Feng and Rithesh R N and Liangwei Yang and Silvio Savarese and Juan Carlos Niebles and Huan Wang and Shelby Heinecke and Caiming Xiong},
booktitle={The Thirty-eight Conference on Neural Information Processing Systems Datasets and Benchmarks Track},
year={2024},
url={https://openreview.net/forum?id=Jfg3vw2bjx}
}

@misc{qian2025toolrlrewardtoollearning,
      title={ToolRL: Reward is All Tool Learning Needs}, 
      author={Cheng Qian and Emre Can Acikgoz and Qi He and Hongru Wang and Xiusi Chen and Dilek Hakkani-Tür and Gokhan Tur and Heng Ji},
      year={2025},
      eprint={2504.13958},
      archivePrefix={arXiv},
      primaryClass={cs.LG},
      url={https://arxiv.org/abs/2504.13958}, 
}

@INPROCEEDINGS{ADC,
  author={Zhang, Wei and Zhang, Yi and Zhu, Li and Jia, Qianghuai and Jiang, Feijun and Guo, Hongcheng and Li, Zhoujun and Zhou, Mengping},
  booktitle={ICASSP 2025 - 2025 IEEE International Conference on Acoustics, Speech and Signal Processing (ICASSP)}, 
  title={ADC: Enhancing Function Calling Via Adversarial Datasets and Code Line-Level Feedback}, 
  year={2025},
  volume={},
  number={},
  pages={1-5},
  doi={10.1109/ICASSP49660.2025.10888405}}

@inproceedings{
wang2025toolgen,
title={ToolGen: Unified Tool Retrieval and Calling via Generation},
author={Renxi Wang and Xudong Han and Lei Ji and Shu Wang and Timothy Baldwin and Haonan Li},
booktitle={The Thirteenth International Conference on Learning Representations},
year={2025},
url={https://openreview.net/forum?id=XLMAMmowdY}
}

@inproceedings{christiano_deep_2017,
	title = {Deep {Reinforcement} {Learning} from {Human} {Preferences}},
	volume = {30},
	url = {https://proceedings.neurips.cc/paper_files/paper/2017/file/d5e2c0adad503c91f91df240d0cd4e49-Paper.pdf},
	booktitle = {Advances in {Neural} {Information} {Processing} {Systems}},
	publisher = {Curran Associates, Inc.},
	author = {Christiano, Paul F and Leike, Jan and Brown, Tom and Martic, Miljan and Legg, Shane and Amodei, Dario},
	editor = {Guyon, I. and Luxburg, U. Von and Bengio, S. and Wallach, H. and Fergus, R. and Vishwanathan, S. and Garnett, R.},
	year = {2017},
}

@inproceedings{
schick2023toolformer,
title={Toolformer: Language Models Can Teach Themselves to Use Tools},
author={Timo Schick and Jane Dwivedi-Yu and Roberto Dessi and Roberta Raileanu and Maria Lomeli and Eric Hambro and Luke Zettlemoyer and Nicola Cancedda and Thomas Scialom},
booktitle={Thirty-seventh Conference on Neural Information Processing Systems},
year={2023},
url={https://openreview.net/forum?id=Yacmpz84TH}
}

@software{text2vec,
  author = {Ming Xu},
  title = {text2vec: A Tool for Text to Vector},
  year = {2023},
  url = {https://github.com/shibing624/text2vec},
}

@inproceedings{
lin2025robust,
title={Robust Function-Calling for On-Device Language Model via Function Masking},
author={Qiqiang Lin and Muning Wen and Qiuying Peng and Guanyu Nie and Junwei Liao and Jun Wang and Xiaoyun Mo and Jiamu Zhou and Cheng Cheng and Yin Zhao and Jun Wang and Weinan Zhang},
booktitle={The Thirteenth International Conference on Learning Representations},
year={2025},
url={https://openreview.net/forum?id=yVQcr4qjD6}
}

@article{goodfellow2014explaining,
  title={Explaining and harnessing adversarial examples},
  author={Goodfellow, Ian J and Shlens, Jonathon and Szegedy, Christian},
  journal={arXiv preprint arXiv:1412.6572},
  year={2014}
}

@inproceedings{patil2025bfcl,
title={The Berkeley Function Calling Leaderboard (BFCL): From Tool Use to Agentic Evaluation of Large Language Models}, 
author={Patil, Shishir G. and Mao, Huanzhi and Cheng-Jie Ji, Charlie and Yan, Fanjia and Suresh, Vishnu and Stoica, Ion and E. Gonzalez, Joseph},
booktitle={Forty-second International Conference on Machine Learning},
year={2025},
}

@misc{goodfellow2014generativeadversarialnetworks,
      title={Generative Adversarial Networks}, 
      author={Ian J. Goodfellow and Jean Pouget-Abadie and Mehdi Mirza and Bing Xu and David Warde-Farley and Sherjil Ozair and Aaron Courville and Yoshua Bengio},
      year={2014},
      eprint={1406.2661},
      archivePrefix={arXiv},
      primaryClass={stat.ML},
      url={https://arxiv.org/abs/1406.2661}, 
}

@misc{zheng2024optimalllmalignmentsusing,
      title={Toward Optimal LLM Alignments Using Two-Player Games}, 
      author={Rui Zheng and Hongyi Guo and Zhihan Liu and Xiaoying Zhang and Yuanshun Yao and Xiaojun Xu and Zhaoran Wang and Zhiheng Xi and Tao Gui and Qi Zhang and Xuanjing Huang and Hang Li and Yang Liu},
      year={2024},
      eprint={2406.10977},
      archivePrefix={arXiv},
      primaryClass={cs.CL},
      url={https://arxiv.org/abs/2406.10977}, 
}

@misc{deng2025duoguardtwoplayerrldrivenframework,
      title={DuoGuard: A Two-Player RL-Driven Framework for Multilingual LLM Guardrails}, 
      author={Yihe Deng and Yu Yang and Junkai Zhang and Wei Wang and Bo Li},
      year={2025},
      eprint={2502.05163},
      archivePrefix={arXiv},
      primaryClass={cs.CL},
      url={https://arxiv.org/abs/2502.05163}, 
}

@article{wang2025function,
  title={Function Calling in Large Language Models: Industrial Practices, Challenges, and Future Directions},
  author={WANG, MAOLIN and ZHANG, YINGYI and PENG, CUNYIN and CHEN, YICHENG and ZHOU, WEI and GU, JINJIE and ZHUANG, CHENYI and GUO, RUOCHENG and YU, BOWEN and WANG, WANYU and others},
  year={2025}
}

@article{qin2024tool,
  title={Tool learning with foundation models},
  author={Qin, Yujia and Hu, Shengding and Lin, Yankai and Chen, Weize and Ding, Ning and Cui, Ganqu and Zeng, Zheni and Zhou, Xuanhe and Huang, Yufei and Xiao, Chaojun and others},
  journal={ACM Computing Surveys},
  volume={57},
  number={4},
  pages={1--40},
  year={2024},
  publisher={ACM New York, NY}
}

@misc{zhong2025complexfuncbenchexploringmultistepconstrained,
      title={ComplexFuncBench: Exploring Multi-Step and Constrained Function Calling under Long-Context Scenario}, 
      author={Lucen Zhong and Zhengxiao Du and Xiaohan Zhang and Haiyi Hu and Jie Tang},
      year={2025},
      eprint={2501.10132},
      archivePrefix={arXiv},
      primaryClass={cs.CL},
      url={https://arxiv.org/abs/2501.10132}, 
}

@inproceedings{
wu2025selfplay,
title={Self-Play Preference Optimization for Language Model Alignment},
author={Yue Wu and Zhiqing Sun and Huizhuo Yuan and Kaixuan Ji and Yiming Yang and Quanquan Gu},
booktitle={The Thirteenth International Conference on Learning Representations},
year={2025},
url={https://openreview.net/forum?id=a3PmRgAB5T}
}

@inproceedings{self-fine,
author = {Chen, Zixiang and Deng, Yihe and Yuan, Huizhuo and Ji, Kaixuan and Gu, Quanquan},
title = {Self-play fine-tuning convertsweak language models to strong language models},
year = {2024},
publisher = {JMLR.org},
booktitle = {Proceedings of the 41st International Conference on Machine Learning},
articleno = {256},
numpages = {22},
location = {Vienna, Austria},
series = {ICML'24}
}

@inproceedings{
cheng2024selfplaying,
title={Self-playing Adversarial Language Game Enhances {LLM} Reasoning},
author={Pengyu Cheng and Tianhao Hu and Han Xu and Zhisong Zhang and Yong Dai and Lei Han and nan du and Xiaolong Li},
booktitle={The Thirty-eighth Annual Conference on Neural Information Processing Systems},
year={2024},
url={https://openreview.net/forum?id=oCGkSH7ys2}
}

@misc{chen2025acebenchwinsmatchpoint,
      title={ACEBench: Who Wins the Match Point in Tool Usage?}, 
      author={Chen Chen and Xinlong Hao and Weiwen Liu and Xu Huang and Xingshan Zeng and Shuai Yu and Dexun Li and Shuai Wang and Weinan Gan and Yuefeng Huang and Wulong Liu and Xinzhi Wang and Defu Lian and Baoqun Yin and Yasheng Wang and Wu Liu},
      year={2025},
      eprint={2501.12851},
      archivePrefix={arXiv},
      primaryClass={cs.CL},
      url={https://arxiv.org/abs/2501.12851}, 
}

@misc{kate2025longfuncevalmeasuringeffectivenesslong,
      title={LongFuncEval: Measuring the effectiveness of long context models for function calling}, 
      author={Kiran Kate and Tejaswini Pedapati and Kinjal Basu and Yara Rizk and Vijil Chenthamarakshan and Subhajit Chaudhury and Mayank Agarwal and Ibrahim Abdelaziz},
      year={2025},
      eprint={2505.10570},
      archivePrefix={arXiv},
      primaryClass={cs.SE},
      url={https://arxiv.org/abs/2505.10570}, 
}

@misc{schulman2017proximalpolicyoptimizationalgorithms,
      title={Proximal Policy Optimization Algorithms}, 
      author={John Schulman and Filip Wolski and Prafulla Dhariwal and Alec Radford and Oleg Klimov},
      year={2017},
      eprint={1707.06347},
      archivePrefix={arXiv},
      primaryClass={cs.LG},
      url={https://arxiv.org/abs/1707.06347}, 
}

@article{zhou2024reflect,
  title={Reflect-rl: Two-player online rl fine-tuning for lms},
  author={Zhou, Runlong and Du, Simon S and Li, Beibin},
  journal={arXiv preprint arXiv:2402.12621},
  year={2024}
}

@article{ma2024coevolving,
  title={Coevolving with the other you: Fine-tuning llm with sequential cooperative multi-agent reinforcement learning},
  author={Ma, Hao and Hu, Tianyi and Pu, Zhiqiang and Boyin, Liu and Ai, Xiaolin and Liang, Yanyan and Chen, Min},
  journal={Advances in Neural Information Processing Systems},
  volume={37},
  pages={15497--15525},
  year={2024}
}

@article{patil2024gorilla,
  title={Gorilla: Large language model connected with massive apis},
  author={Patil, Shishir G and Zhang, Tianjun and Wang, Xin and Gonzalez, Joseph E},
  journal={Advances in Neural Information Processing Systems},
  volume={37},
  pages={126544--126565},
  year={2024}
}

@misc{zhao2025absolutezeroreinforcedselfplay,
      title={Absolute Zero: Reinforced Self-play Reasoning with Zero Data}, 
      author={Andrew Zhao and Yiran Wu and Yang Yue and Tong Wu and Quentin Xu and Yang Yue and Matthieu Lin and Shenzhi Wang and Qingyun Wu and Zilong Zheng and Gao Huang},
      year={2025},
      eprint={2505.03335},
      archivePrefix={arXiv},
      primaryClass={cs.LG},
      url={https://arxiv.org/abs/2505.03335}, 
}

@inproceedings{yao2023react,
  title={React: Synergizing reasoning and acting in language models},
  author={Yao, Shunyu and Zhao, Jeffrey and Yu, Dian and Du, Nan and Shafran, Izhak and Narasimhan, Karthik and Cao, Yuan},
  booktitle={International Conference on Learning Representations (ICLR)},
  year={2023}
}

@article{sheng2024hybridflow,
  title   = {HybridFlow: A Flexible and Efficient RLHF Framework},
  author  = {Guangming Sheng and Chi Zhang and Zilingfeng Ye and Xibin Wu and Wang Zhang and Ru Zhang and Yanghua Peng and Haibin Lin and Chuan Wu},
  year    = {2024},
  journal = {arXiv preprint arXiv: 2409.19256}
}

\end{document}